\documentclass[trackchanges]{aastex701}

\usepackage[labelformat=simple]{subfig}
\usepackage{booktabs}     
\usepackage{threeparttable} 
\usepackage{tabularx}      
\usepackage{array}

\begin{document}

\title{Artificial Satellite Trails Detection Using U-Net Deep Neural Network and Line Segment Detector Algorithm}

\author{Xiaohan Chen}
\affiliation{School of Physics and Astronomy, China West Normal University, Nanchong 637002, People's Republic of China}
\affiliation{National Astronomical Observatories, Chinese Academy of Sciences, Beijing 100101,  People's Republic of China}
\email{xhchen@bao.ac.cn}

\author{Hongrui Gu}
\affiliation{School of Astronomy and Space Science, University of Chinese Academy of Sciences, Beijing 100049, People's Republic of China}
\affiliation{National Astronomical Observatories, Chinese Academy of Sciences, Beijing 100101,  People's Republic of China}
\email{guhr@bao.ac.cn}

\author{Cunshi Wang}
\affiliation{School of Astronomy and Space Science, University of Chinese Academy of Sciences, Beijing 100049, People's Republic of China}
\affiliation{National Astronomical Observatories, Chinese Academy of Sciences, Beijing 100101,  People's Republic of China}
\email{wangcunshi@nao.cas.cn}

\author{Haiyang Mu}
\affiliation{National Astronomical Observatories, Chinese Academy of Sciences, Beijing 100101,  People's Republic of China}
\email{hymu@bao.ac.cn}

\author{Jie Zheng}
\affiliation{National Astronomical Observatories, Chinese Academy of Sciences, Beijing 100101,  People's Republic of China}
\email{jiezheng@nao.cas.cn}

\author{Junju Du}
\affiliation{Shandong Provincial Key Laboratory of Optical Astronomy and Solar-Terrestrial Environment, School of Space Science and Technology, Institute of Space Sciences, Shandong University, Weihai, 264209, Shandong, People’s Republic of China}
\email{dujunju@sdu.edu.cn}

\author{Jing Ren}
\affiliation{National Astronomical Observatories, Chinese Academy of Sciences, Beijing 100101,  People's Republic of China}
\affiliation{School of Astronomy and Space Science, University of Chinese Academy of Sciences, Beijing 100049, People's Republic of China}
\email{rj@bao.ac.cn}

\author{Zhou Fan$^{\dagger}$}
\affiliation{National Astronomical Observatories, Chinese Academy of Sciences, Beijing 100101,  People's Republic of China}
\affiliation{School of Astronomy and Space Science, University of Chinese Academy of Sciences, Beijing 100049, People's Republic of China}
\email{zfan@nao.cas.cn}
\altaffiliation{Also corresponding author} 

\author{Jing Li$^{\ast}$}
\affiliation{School of Physics and Astronomy, China West Normal University, Nanchong 637002, People's Republic of China}
\email{lijing@bao.ac.cn}
\altaffiliation{Corresponding author} 


\begin{abstract}

     With the rapid increase in the number of artificial satellites, astronomical imaging is experiencing growing interference. 
     When these satellites reflect sunlight, they produce streak-like artifacts in photometry images. 
     Such satellite trails can introduce false sources and cause significant photometric errors. 
     As a result, accurately identifying the positions of satellite trails in observational data has become essential. 
     In this work, we propose a satellite trail detection model that combines the U-Net deep neural network for image segmentation with the Line Segment Detector (LSD) algorithm. 
     The model is trained on 375 simulated images of satellite trails, generated using data from the Mini-SiTian Array. 
     Experimental results show that for trails with a signal-to-noise ratio (SNR) greater than 3, the detection rate exceeds 99\%. 
     Additionally, when applied to real observational data from the Mini-SiTian Array, the model achieves a recall of 79.57\% and a precision of 74.56\%.

     \vspace{1em} 
     \noindent
     $^{\ast}$Corresponding author: Jing Li (lijing@bao.ac.cn) \\
     $^{\dagger}$Also corresponding author: Zhou Fan (zfan@nao.cas.cn)

\end{abstract}



\section{Introduction} \label{sect:intro}

The number of artificial satellites in Low Earth Orbit (LEO) has increased 
dramatically in recent years, driven in large part by the deployment of 
large-scale satellite constellations such as Starlink 
\citep{shaengchart2024spacex, walker2021starlink}, 
OneWeb \citep{10551683}, Kuiper Systems \citep{2022NewSc.254...15M}, 
and the Qianfan Constellation \citep{2025JPhCS3073a2006Y}. 
According to the UCS Satellite Database, there are currently over 7,560 operational satellites in Earth's orbit \citep{ucs_sat_db_2023}.
These large-scale satellite constellations are expected to at least double the current satellite population.
Projections indicate that by 2030, the number of LEO satellites 
dedicated to communication alone will exceed 100,000 \citep{Lawrence2022}.

These satellite constellations pose a significant threat to astronomical 
research \citep{2023NatAs...7..252B}. 
Artificial satellites often reflect sunlight, creating streak-like trails in observational images. 
These trails contaminate imaging and spectral data, introduce false sources, cause photometric errors, and generate false spectral signals. 
As a result, many observations affected by satellite contamination become unsuitable for scientific analysis, 
leading to substantial portions of research budgets being spent 
on mitigating these impacts \citep{2023NatAs...7..262K,2023MNRAS.525L..60K}.
To assess these effects, \cite{2024A&A...687A.135H} developed a satellite probability distribution model, which evaluates the extent of satellite interference on telescopes. Their findings indicate that ground-based telescopes are particularly affected in regions near the equator, with latitudes around ±50 degrees and ±80 degrees experiencing the highest levels of interference from LEO satellites.

Large-scale time-domain surveys are becoming increasingly prevalent, and the impact of artificial satellite trails on astronomical observations is expected to grow correspondingly. 
For instance, the Starlink satellite constellation significantly affects both 
the Large Synoptic Survey Telescope (LSST) \citep{2019ApJ...873..111I} and 
the Zwicky Transient Facility (ZTF) \citep{Bellm_2018}. 
During LSST's twilight and low-elevation observations, 
satellite trails appear in approximately one out of every nine exposures, and only about half of these trails have brightness levels 
within the correctable range \citep{2024arXiv240311118L}. 
ZTF has experienced similar interference during twilight, with the fraction of affected images rising dramatically from 0.5\% in 2019 to 18\% in 2021. 
Projections indicate that if the number of satellites reaches 10,000, nearly all ZTF twilight images could be contaminated 
by satellite trails \citep{2022ApJ...924L..30M}. 
Similarly, the SiTian project \citep{2021AnABC..93..628L}—a ground-based, all-sky optical survey developed by the Chinese Academy of Sciences—is vulnerable to satellite interference.
SiTian consists of over 60 globally distributed 1-meter-class telescopes that form a unified network aimed at detecting Gravitational Wave Electromagnetic Counterpart (GWEM), 
studying early supernova light curves, exploring intermediate-luminosity transients, investigating Tidal Disruption Events (TDE), 
identifying optical counterparts of gamma-ray bursts (GRBs) and fast radio bursts (FRBs), and monitoring small solar system bodies such as asteroids and comets through high-cadence (sub-hourly) observations.
Due to their high cadence, these large-scale time-domain surveys produce a vast number of images, increasing the probability of contamination by satellite trails. Additionally, their wide fields of view make it more probable that multiple satellite trails will appear in each frame. As the satellite population continues to grow, the resulting contamination may degrade the quality of long-term observational data. Satellite trails also complicate image preprocessing workflows. Therefore, to minimize the impact of satellite trails on photometric analysis, it is crucial to identify their positions and flag the sources they affect.

\begin{figure}
\centering
\includegraphics[width=\textwidth, angle=0]{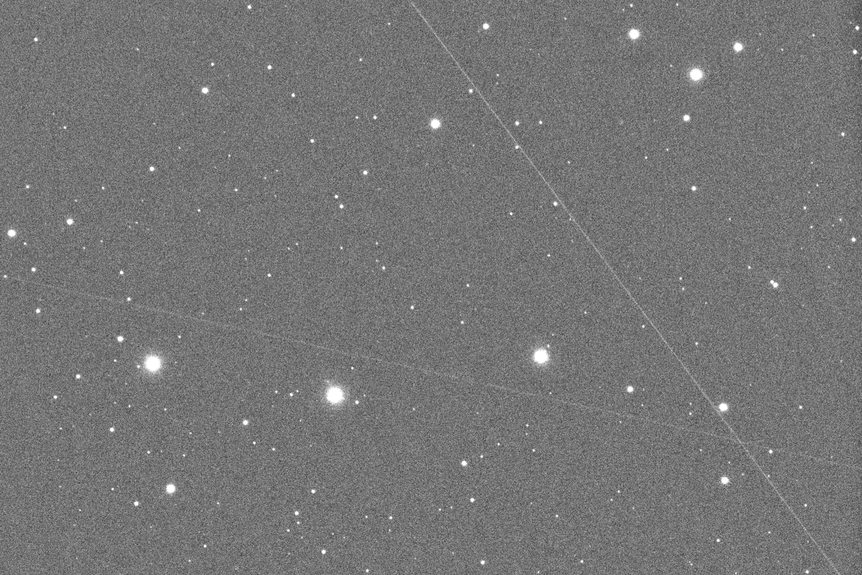}
\caption{\small An example of observational data from the Mini-SiTian Array at Xinglong Observatory, showing two artificial satellite trails intersecting the field of view.}
\label{sub_mst}
\end{figure}

Detection methods for stripe-shaped trails in images can generally be categorized 
into four main families \citep{2024PrA....42..473W}. 
The first category includes traditional digital image processing techniques. 
These methods are fast, computationally efficient, and relatively stable. 
However, their performance is often compromised by complex or variable backgrounds, 
leading to increased rates of false positives and missed detections. 
A prominent example in this category is the Hough Transform \citep{Hough1962}, 
a classical technique for detecting straight lines in images. 
It has been widely adopted in astronomical applications, 
including its implementation in IRAF 
for line detection \citep{1999ASPC..172..349C}. 
Over time, the Hough Transform has also been extended to detect a broader range of patterns 
beyond straight lines \citep{1981PatRe..13..111B}.
StreakDet is also widely used among the community as an automated processing pipeline for optical observations of moving objects \citep{2014acm..conf..570V}. It detects low-SNR streaks without requiring prior velocity information by applying local grayscale mean-difference evaluation and thresholding, and further enables orbital validation of the detected streaks.

The second and third categories are matched filtering methods and intensity distribution methods, respectively. 
In practice, 
matched filtering can be considered a specific subtype of intensity distribution methods \citep{2024PrA....42..473W}. 
These approaches exploit the differences in intensity profiles between trails and background noise to effectively distinguish and detect the trails.
\cite{2018AJ....156..229N} used the Radon transform to identify asteroid streaks, satellite streaks and diffraction spikes, among others, and their method was very powerful for detecting faint streaks.
\cite{2020AdSpR..65..364V} detects faint object streaks by convolving the image with filters matching potential streak geometries and orientations, using background statistics to set thresholds, effectively identifying long streaks even at SNRs as low as 0.5.
However, the detection time of this method is significantly affected by the image size, with images as large as 4000$\times$4000 pixels even requiring 2-3 hours.
While these methods have achieved some success, they still pose a significant challenge when dealing with issues like complex backgrounds and also require computationally intensive convolution with multiple templates across angles and scales \citep{2024A&A...692A.199S,2024PrA....42..473W}.

The final category comprises machine learning-based methods. 
These approaches rely on training models with large volumes of labeled data to automatically extract features 
and make predictions on new, unseen data. 
Among them, convolutional neural networks (CNNs) have proven 
particularly effective for image analysis tasks \citep{2015Natur.521..436L}. 
CNNs learn to identify image features through a hierarchical architecture that includes convolutional layers, 
pooling layers, and fully connected layers, enabling robust recognition and processing of complex visual patterns. 
\cite{2024AcAau.225..978M} demonstrated the effectiveness of combining 
the You Only Look Once (YOLO) algorithm \citep{7780460} 
with the U-Net architecture \citep{10.1007/978-3-319-24574-4_28} to detect streak-like patterns associated with Resident Space Objects (RSOs), achieving strong performance. 
\cite{2025PASP..137e4503I,2025A&A...694A..49I} also used U-Net for asteroid streaks detection, which are typically shorter than the satellite streaks, but share many similar features.
In comparison, artificial satellite trails often present an even greater challenge. 
These trails frequently appear as bright, continuous streaks that span the entire field of view during long exposures—often lasting a minute or more—and can contaminate a significant portion of the data within a single frame.
However, researchers have demonstrated that U-Net also performs exceptionally well in the task of detecting artificial satellite trails. 
\cite{article} used U-Net to effectively filter out artifacts caused by LEO satellites in astronomical observations, 
which outperforms existing methods in overall accuracy with significantly reduced computational time, 
demonstrating the promising potential of applying previously under-explored deep learning techniques to satellite artifact removal.

Machine learning methods excel at learning target features from diverse datasets, 
but they typically require large amounts of labeled training data and considerable computational resources. 
In contrast, classical line detection algorithms—while efficient and accurate 
in relatively simple backgrounds—often struggle in more complex environments, 
such as astronomical images, where background noise and variability can hinder performance \citep{2024PrA....42..473W}.

Notably, the work of \cite{Teplyakov_2022} demonstrated that incorporating a lightweight convolutional neural network (CNN) into a classical line segment detection framework significantly improves both speed and accuracy. 
Their hybrid model achieved detection accuracies of 77.5\% 
on the ``Wireframe" dataset \citep{5b5f04115d73410982dc64235ddaf300} 
and 64.6\% on the ``York Urban" dataset \citep{2008Efficient}, both of which primarily consist of architectural and urban scenes. 
Meanwhile, \cite{2024A&A...692A.199S} have confirmed that such an approach is equally efficient in applications involving astronomical images.
They combined U-Net and the Hough transform to effectively detect and analyze satellite trails in ground-based astronomical observations. 
However, the performance of their method on satellite trails with varying SNR, particularly those with low SNRs, remains to be explored.
Moreover, the Hough transform is known to be sensitive to parameter selection, which can complicate its application in diverse observational conditions.

Motivated by these findings, this study aims to develop a lightweight and efficient method for detecting artificial satellite trails in optical astronomical images, 
using data from the Mini-SiTian Array—a pathfinder for the SiTian project. 
Our approach combines a lightweight deep learning model, U-Net, with the Line Segment Detector (LSD), 
a classical line detection algorithm known for its adaptivity. 
By leveraging the strengths of both methodologies, our goal is to achieve high detection capability across a wide range of SNRs, including challenging low SNR conditions.
The model is trained on simulated data and its performance is evaluated on both simulated and real observational data. 
The remainder of this paper is structured as follows: Section \ref{sect:data} describes the simulated and real observational datasets obtained from the Mini-SiTian Array. Section \ref{sect:method} outlines the methodology used to construct the satellite trail detection model. Section \ref{sect:results} presents the experimental results along with relevant analysis and discussion. Finally, Section \ref{sect:conclusion} summarizes the study and outlines potential directions for future work.

\section{Data}
\label{sect:data}

To construct the training and test datasets, we used observational data from the Mini-SiTian Array 
that did not contain artificial satellite trails and simulated the presence of trails for model development. 
For model evaluation, we selected one week of real observational data—from December 18 to December 24, 2024—comprising a total of 714 valid images. 
To facilitate both automated model testing and manual inspection of satellite trails, each full-size image was divided into 20 overlapping sub-images, resulting in a dataset of 14,280 sub-images. Among these, 1,229 sub-images contained one or more satellite trails, with a total of 1,488 trails identified across the dataset.

\subsection{The Mini-SiTian Data}
\label{subsec:mst_data}

The Mini-SiTian project \citep{2025RAA....25d4005H}, a pathfinder of SiTian, is located at the Xinglong Observatory. 
It serves as a testbed for validating the feasibility of SiTian’s control systems, data processing pipelines, and observational strategies, 
while also supporting a range of preliminary research efforts \citep{2025RAA....25d4001H}. 
The system comprises three 30-cm-aperture telescopes, each equipped with g, r, and i filters. 
Each telescope features a wide field of view of $2.29^{\circ} \times 1.53^{\circ}$ and uses a 9K $\times$ 6K scientific CMOS detector. 
With an exposure time of 300 seconds, 
the system achieves 5-sigma limiting magnitudes of 19.5 in the g-band and 19.0 in the r-band \citep{2025RAA....25d4005H}. 
The large field of view of the Mini-SiTian telescopes increases the likelihood of capturing artificial satellite trails in a single exposure. 
However, the relatively small aperture size (30 cm) results in a lower SNR, particularly for faint trails, thereby increasing both the level of interference and the difficulty of detection (as illustrated in Figure \ref{sub_mst}).

\subsection{Data Simulation}
\label{subsec:sim_data}

\begin{figure*}
\centering
\subfloat[]{
     \includegraphics[width=0.45\textwidth]{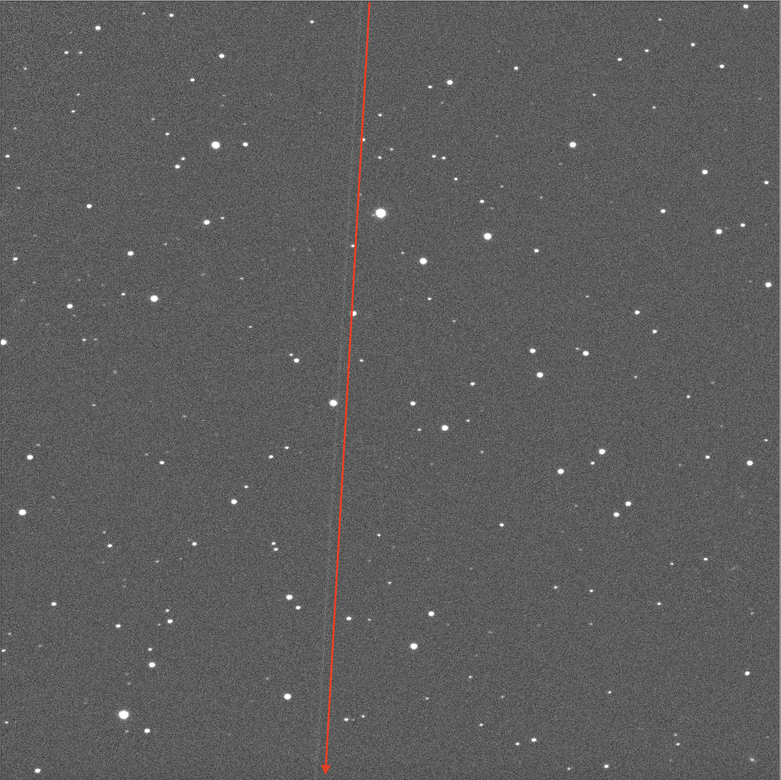}
     \label{fig:snr3}
}
\hfill
\subfloat[]{
     \includegraphics[width=0.45\textwidth]{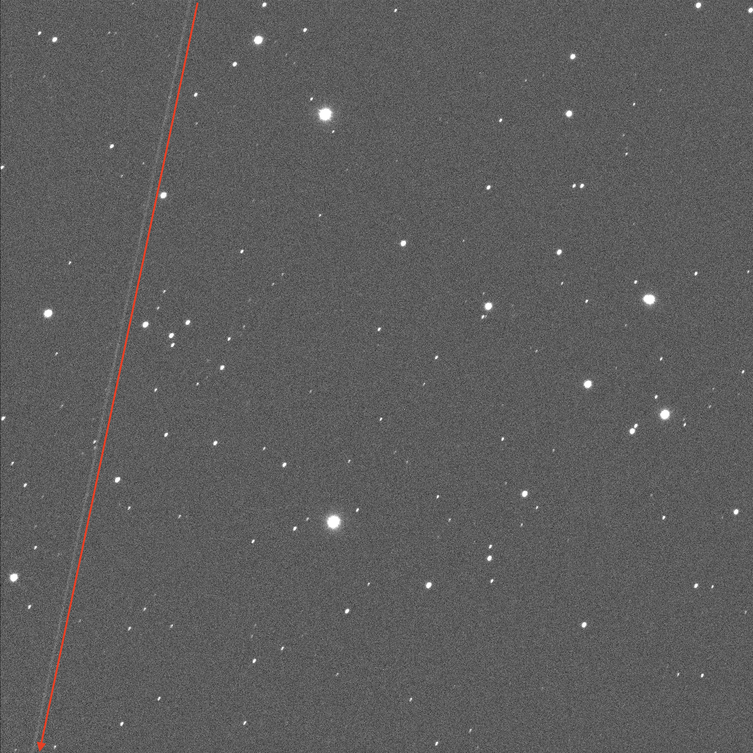}
     \label{fig:snr5}
}

\vspace{1cm}

\subfloat[]{
     \includegraphics[width=0.45\textwidth]{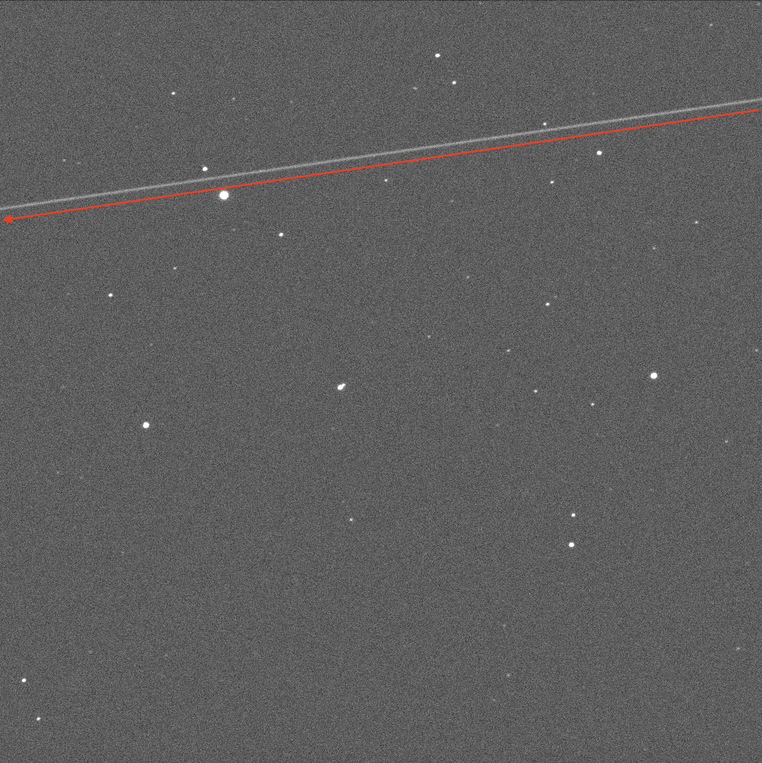}
     \label{fig:snr13}
}
\hfill
\subfloat[]{
     \includegraphics[width=0.45\textwidth]{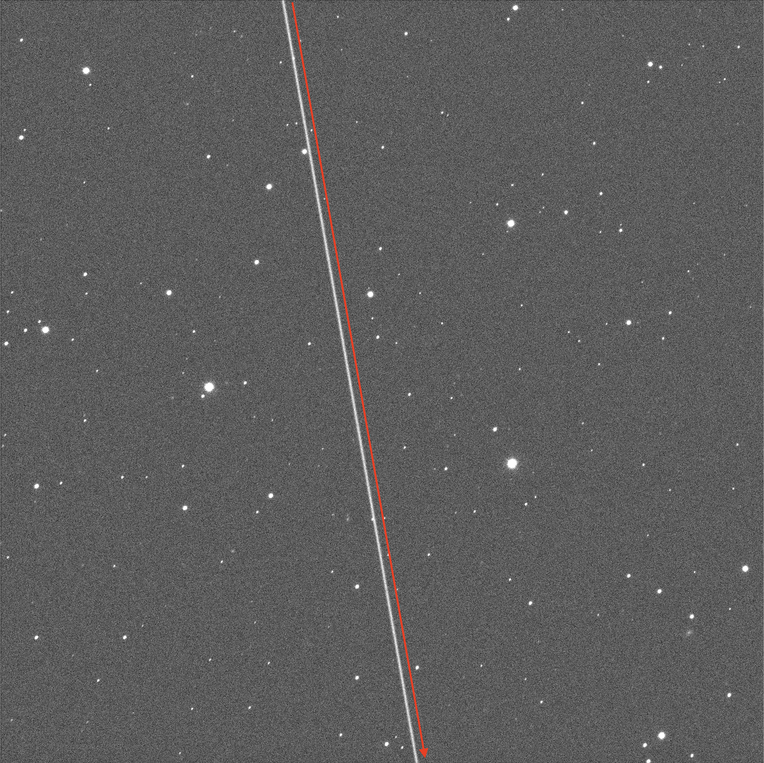}
     \label{fig:snr22}
}

\caption{\small Panels (a)-(d) display simulated satellite trails with increasing SNR: (a) SNR = 3; (b) SNR = 5; (c) SNR = 15; (d) SNR = 30. 
The white lines represent the simulated satellite trails, while the solid red lines are supplementary annotations manually added for visualization purposes to highlight the positions of these simulated streaks. Note that these red markers are not part of the training labels or the actual training images; they were introduced during post-processing solely to aid interpretation in this figure.}
\label{dataset}
\end{figure*}

Deep learning methods typically require large volumes of training data to achieve robust performance. To ensure that our model was trained on a sufficiently diverse set of artificial satellite trail images with a wide range of SNRs, we constructed a comprehensive simulated dataset.

To construct the simulated dataset, we selected background images from the Mini-SiTian Array at Xinglong Observatory that were free of artificial satellite trails. 
A two-dimensional Gaussian fitting was applied to the stars in these images to characterize their Point Spread Function (PSF). 
Since atmospheric seeing can be assumed to remain approximately constant over the short timescale during which a satellite crosses the field of view, 
we modeled satellite trails using the same PSF as that of the stars. 
Artificial satellites reflect sunlight, and due to variations in solar illumination angles and changes in satellite orientation, 
their trails often exhibit periodic fluctuations in brightness. 
However, our model does not explicitly represent the brightness fluctuations.
U-Net consists of a contracting path for feature extraction and an expansive path for precise localization, 
enabling effective segmentation of features across multiple spatial scales \citep{10.1007/978-3-319-24574-4_28}.
To enhance learning efficiency and strengthen the robustness of the detection pipeline, we prioritized generalized structural features over finely rendered
brightness variations. This approach encourages the model to focus on recognizing continuous linear structures as a class, 
rather than overfitting to specific and inconsistent luminosity patterns.
To generate realistic trail patterns, we randomly drew straight lines across the background images. 
These lines were assigned Gaussian profiles in the width direction and uniformly distributed lengths. 
Poisson noise was added to simulate observational conditions, and the trails were overlaid onto the images. 
Corresponding binary masks were generated as ground-truth labels for supervised learning. We then calculated the SNR for each trail. Since the trail length affects SNR, 
we followed the aperture photometry method used in IRAF and adopted a square aperture with a side length of three times the full width at half maximum (FWHM) to compute the SNR. 
The SNR is calculated using the following equation \citep{1981SPIE..290...28M,1992ASPC...23..130G}:
\begin{equation}
  \frac{S}{N} = \frac{S_\star}{\sqrt{S_\star + n_{\text{p}}\cdot\bigl(1+\frac{n_{\text{p}}}{n_{\text{s}}}\bigr)\cdot(S_S+t\cdot dc+R^2+G^2\sigma_f^2)}}
\end{equation}
\begin{equation}
  \sigma_f^2 = \int_{-1/2}^{1/2}f^2\mathrm{d}f = 0.083
\end{equation}
where $S_\star$ is the number of signal electrons, 
$S_S$ is the average number of sky background electrons per pixel, 
$n_{\text{p}}$ is the number of pixels used to measure the signal, 
$n_{\text{s}}$ is the number of pixels used to estimate the background, 
$t$ is the exposure time, $dc$ is the dark current, $R$ is the readout noise, $G$ is the gain, 
and $\sigma_f$ is the quantization error, which is modeled by an independent uniform random variable ranging from -0.5 to 0.5. 
Although individual points with low SNR may be indistinguishable from noise, the human visual system is adept at detecting continuous patterns through mechanisms such as lateral inhibition \citep{10.1001/archneur.1966.00470070121025}. 
Consequently, satellite trails—despite having similar SNRs to point sources—are more perceptible due to their linear and coherent structure.

Figure \ref{dataset} presents four simulated images containing artificial satellite trails with varying SNRs. 
To enhance the diversity and robustness of our dataset, we selected several distinct background images and generated multiple simulations for each, incorporating trails with different SNR levels. As a result, the final dataset includes a wide range of background conditions. Furthermore, to ensure the generalization capability of our model, we constructed the training and test sets using different sets of background images.

\begin{figure}
\centering
\includegraphics[width=\textwidth, angle=0]{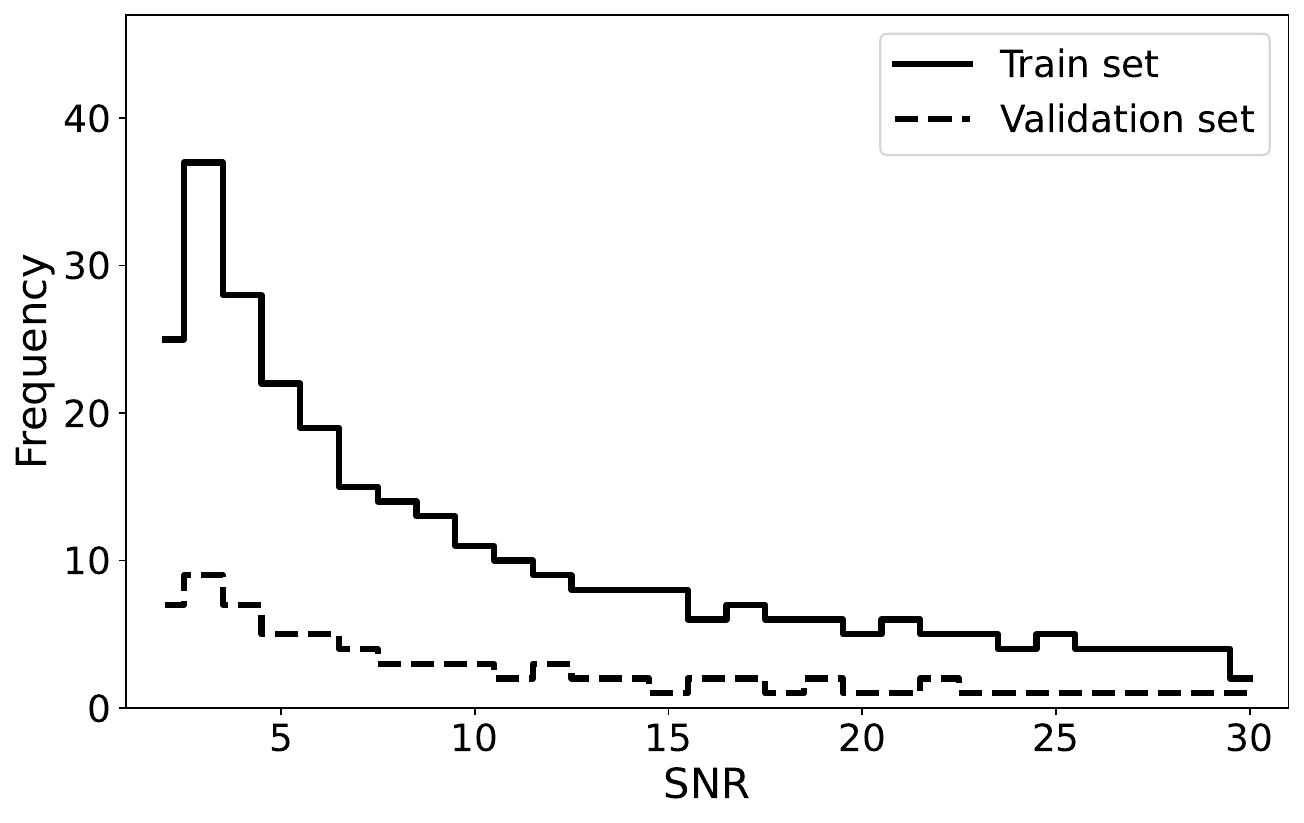}
\caption{\small SNR distribution in the training and validation sets. The training set comprises 300 images, while the validation set includes 75 images.}
\label{snr_distribution}
\end{figure}

\section{Method}
\label{sect:method}

In this study, we developed a model for detecting artificial satellite trails in astronomical images by combining the U-Net deep learning architecture \citep{10.1007/978-3-319-24574-4_28} with the Line Segment Detector (LSD) algorithm \citep{4731268, ipol.2012.gjmr-lsd}.

\subsection{Image Segmentation}
\subsubsection{U-Net}

Deep convolutional neural networks have demonstrated outstanding performance across a wide range of visual recognition tasks, 
leading to their widespread adoption 
in various applications \citep{2015Natur.521..436L,Jiao_2019,8627998,https://doi.org/10.13140/rg.2.2.24831.10403}. 
However, a key limitation of these models is their substantial demand for large-scale annotated datasets and computational resources. 
Many state-of-the-art networks consist of numerous layers and contain millions of parameters. 
Training such deep architectures typically requires thousands of labeled samples and significant computational power, posing challenges for applications with limited data 
or resource constraints \citep{Gu_Ko_Go_Lee_Lee_Shin_2022}.

U-Net \citep{10.1007/978-3-319-24574-4_28} is a convolutional neural network architecture originally proposed for biomedical image segmentation. It features a symmetric encoder–decoder structure, consisting of a contracting path to capture context and an expansive path to enable precise localization. A key advantage of U-Net is its ability to achieve strong performance with relatively small training datasets, aided by extensive data augmentation techniques such as elastic deformations, rotations, and translations. 
As a lightweight network, U-Net is computationally efficient and well-suited for tasks where training data and resources are limited. 
Moreover, its effectiveness in real-time extraction of linear features, such as trails in optical images, has been demonstrated in multiple studies \citep{2022AsDyn...6..205D,9771446}.

\subsubsection{Model Training}

We controlled the SNR of the simulated satellite trails by adjusting the peak pixel values along the width direction. 
To enhance the model’s ability to detect faint and weak trails—common in images from the Mini-SiTian Array—we deliberately increased the proportion of low-SNR samples in the dataset. 
Ultimately, the training dataset comprised 375 simulated satellite trail images with SNRs ranging from 2 to 30. 
These were divided into training and validation sets in an 8:2 ratio, as illustrated in Figure \ref{snr_distribution}. 

\begin{figure}
\centering
\includegraphics[width=\textwidth, angle=0]{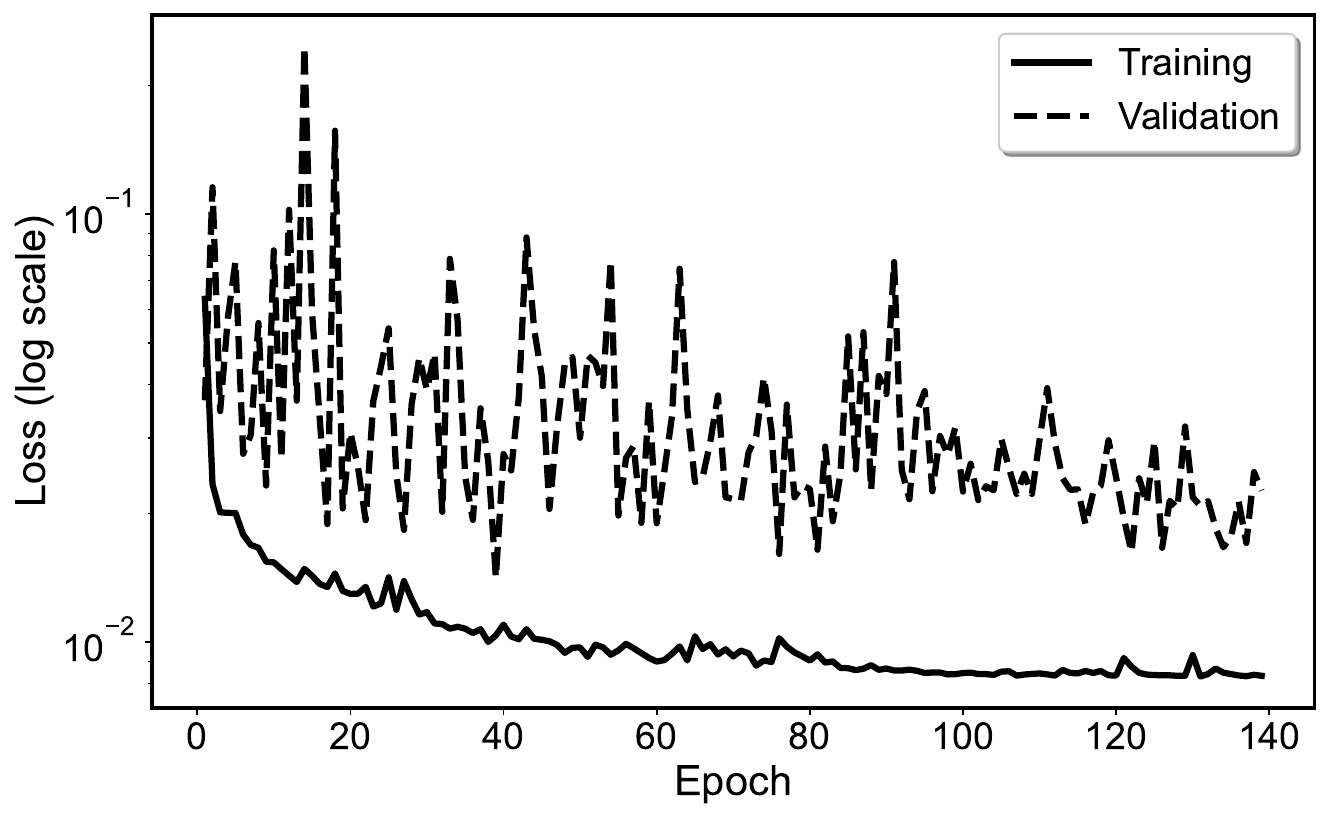}
\caption{\small Total loss versus training epoch of image segmentation network.}
\label{loss}
\end{figure}

During model training, we used a batch size of 2 and employed the Adam optimizer for parameter tuning. 
The initial learning rate was set to 0.01. After 139 epochs, completed over 15.85 hours on a CPU (Apple M2 Pro), 
the model loss stabilized and showed good convergence, as shown in the training loss curve in Figure \ref{loss}. 
This rapid convergence can be attributed to the U-Net architecture’s ability to efficiently capture the core features of satellite streaks, together with the Adam optimizer’s adaptive learning rate, which facilitates effective parameter updates in the early training phase and leads to a swift reduction in loss.
The segmentation of a 256$\times$256 image takes approximately 0.742 seconds.

\subsection{Line Detection}

In cases involving complex backgrounds or faint trails, the U-Net segmentation may generate not only the desired linear features but also various spurious artifacts. 
To enhance the accuracy of detection under such conditions, we apply a classical line detection algorithm to the segmentation output. 
Specifically, we employ the LSD algorithm—a well-established method for detecting straight lines—to refine the segmentation maps generated by U-Net.

The LSD is a linear-time algorithm capable of achieving subpixel accuracy by computing gradient magnitudes and directions across all pixels in an image \citep{4731268, ipol.2012.gjmr-lsd}. 
Both LSD and the Hough Transform are classical line detection algorithms. LSD is computationally efficient, requires no manual parameter tuning, achieves pixel-level accuracy with explicit width and endpoints, whereas the Hough Transform depends on parameter settings and primarily detects infinitely long lines \citep{Hough1962}.
Designed for fully automatic image analysis, LSD internally presets parameters such as the gradient threshold ($\rho$), angle tolerance ($\tau$), and precision ($p$), making it broadly applicable across diverse images without additional adjustment.
The original LSD algorithm detects all line segments in an image, regardless of their length. 
However, we observed that the U-Net segmentation maps introduce some short and fragmented segments due to background interference. 
To mitigate this, we incorporated a length threshold during the LSD post-processing stage to filter out these noise-induced short segments.
Although the large Mini-SiTian images were divided into smaller sub-images for processing, 
the satellite trails remain highly conspicuous within these sub-images—often spanning a significant portion of the sub-image, 
similar to their appearance in the full-scale original images. 
Consequently, even in sub-images, genuine satellite trails are distinctly longer than the short noise segments caused by background artifacts, 
allowing the length threshold to effectively preserve real trails while suppressing noise.
When applied to the segmentation maps produced by U-Net, LSD effectively distinguishes satellite trails from spurious artifacts, as the latter typically consist of short, irregular line segments. 
By enforcing a minimum line-length threshold, we can efficiently suppress non-trail features, thereby enhancing detection accuracy. 
Compared to other classical line detection methods like the Hough Transform—which often necessitate manual tuning for each image—LSD offers superior automation and efficiency. 
Furthermore, LSD’s ability to detect line segments with subpixel precision makes it well-suited for applications requiring high geometric accuracy. Previous work has also demonstrated that integrating lightweight CNNs with classical line detectors can yield strong performance in line segmentation tasks \citep{Teplyakov_2022}. In our implementation, LSD processes a 256$\times$256 U-Net segmentation map in just 0.007 seconds, highlighting its computational efficiency.

\section{Results and Discussion}
\label{sect:results}

\subsection{Test on Simulated Data}
\label{subsec:simulated_result}

To evaluate the performance and generalization capability of the model under varying SNR conditions, 
we constructed a test set consisting of 1,256 simulated images with SNR values ranging from 0 to 30. 
This range was chosen to reflect the wide variability of SNR levels observed in real satellite trail data. 
In practical applications, satellite trails can exhibit a broad spectrum of brightness and noise characteristics, making it essential to assess the model’s effectiveness across diverse SNR scenarios. By incorporating this wide SNR range, we aim to comprehensively evaluate the model’s robustness, adaptability, and overall detection accuracy under both high- and low-noise conditions.

\begin{figure}
\centering
\includegraphics[width=\textwidth, angle=0]{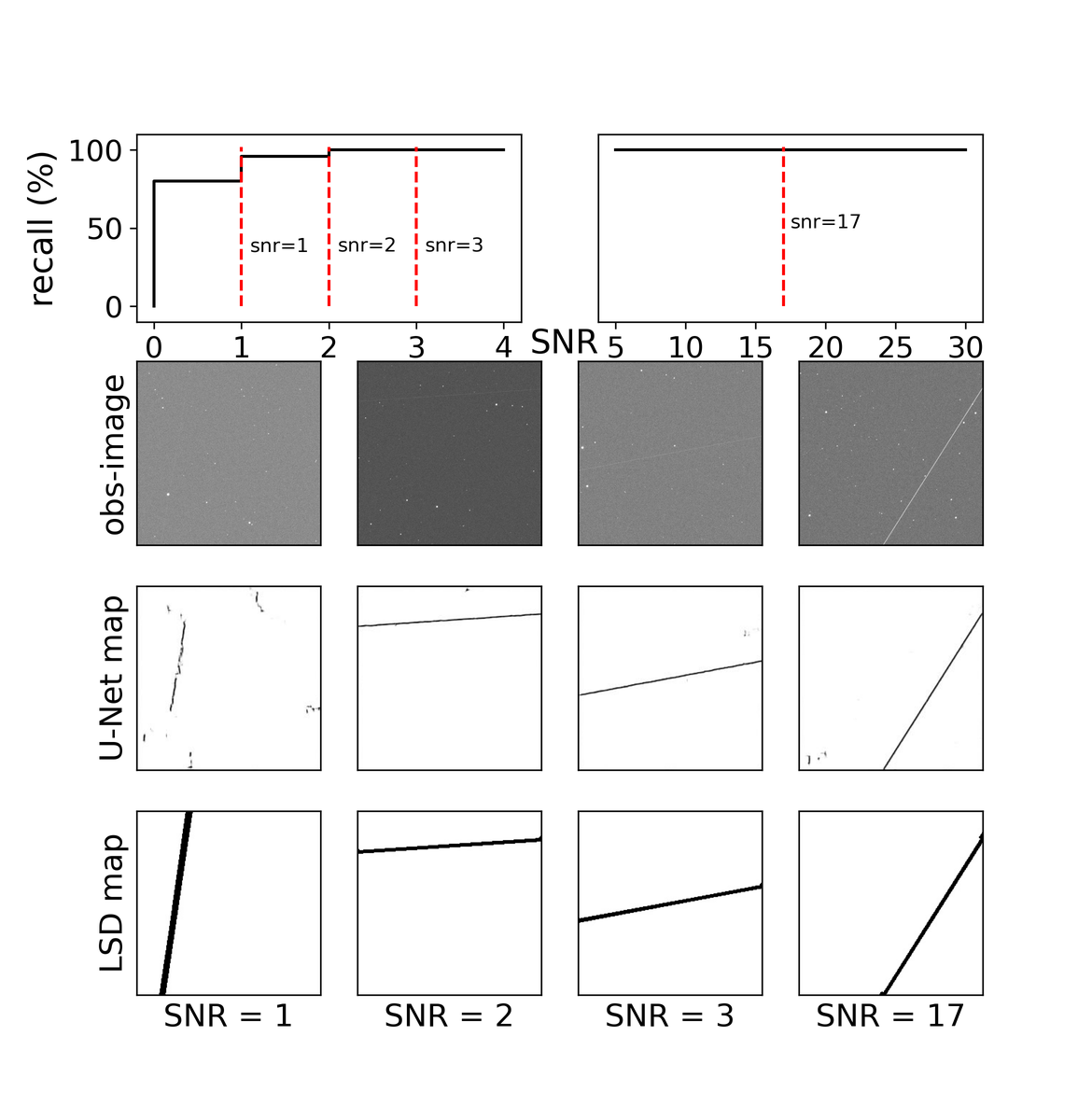}
\caption{\small Detection rate versus SNR of satellite trails. This figure illustrates the model’s detection performance across a range of SNRs, including both faint trails (SNR = 1, 2, 3, etc.) and more prominent trails (e.g., SNR = 17), demonstrating the model’s robustness under varying noise conditions.}
\label{fig:rate}
\end{figure}

Figure \ref{fig:rate} presents the correct detection rate after applying line detection to the segmentation results. 
For images with a SNR greater than 3, the detection rate exceeds 99\%. 
When using U-Net alone for segmentation, background noise or low-SNR trails can sometimes result in additional non-trail patterns, 
which may lead to confusion in identifying artificial satellite trails. 
As a result, the segmentation map may appear less intuitive and more prone to false positives. 
By applying the LSD to the U-Net segmentation output, we can effectively refine the detected features—isolating the satellite trails and suppressing spurious patterns—thereby improving the accuracy and interpretability of the results.

\subsection{Test on Mini-SiTian Data}
\label{subsec:real_result}

\begin{figure*}
\centering
\subfloat[Original image\label{seg_compare1}]{
     \includegraphics[width=0.32\textwidth]{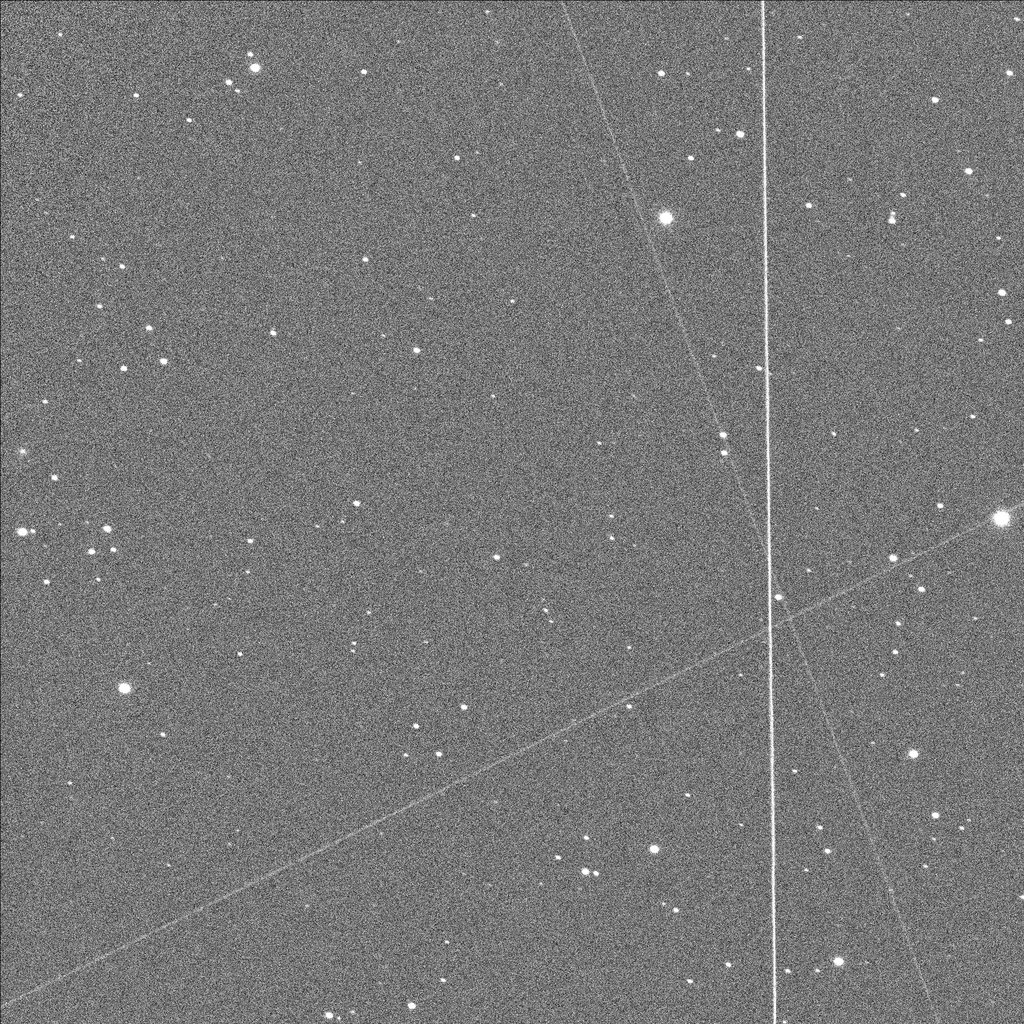}
}
\hfill
\subfloat[U-Net segmentation\label{seg_compare2}]{
     \includegraphics[width=0.32\textwidth]{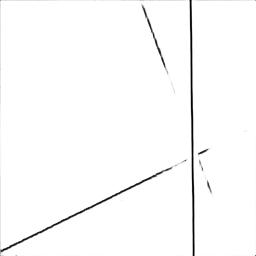}
}
\hfill
\subfloat[LSD processed\label{seg_compare3}]{
     \includegraphics[width=0.32\textwidth]{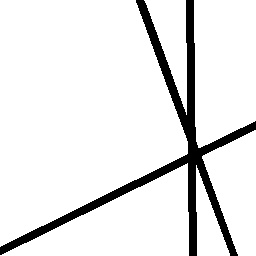}
}

\vspace{0.5cm}

\subfloat[Original image\label{seg_compare21}]{
     \includegraphics[width=0.32\textwidth]{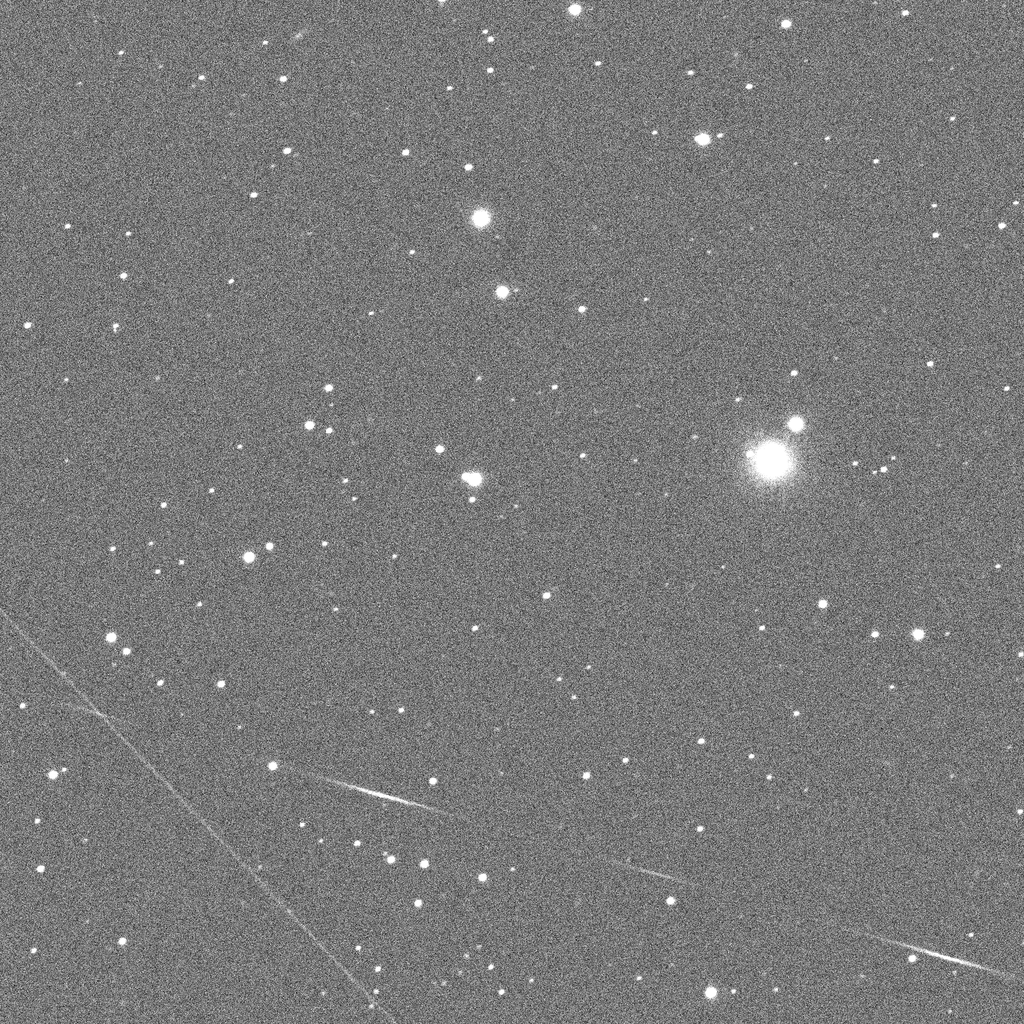}
}
\hfill
\subfloat[U-Net segmentation\label{seg_compare22}]{
     \includegraphics[width=0.32\textwidth]{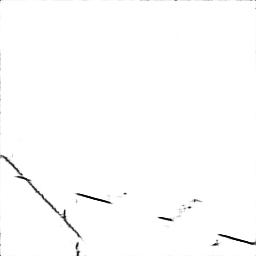}
}
\hfill
\subfloat[LSD processed\label{seg_compare23}]{
     \includegraphics[width=0.32\textwidth]{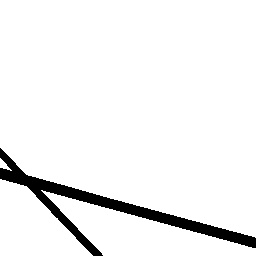}
}

\caption{\small Image segmentation results from real observational data collected by the Mini-SiTian Array No. 3 telescope at the Xinglong Observatory. 
(a) and (d) show the original images; 
(b) and (e) display the segmentation maps generated by U-Net; 
(c) and (f) present the results after applying the LSD algorithm to the U-Net segmentation maps.}
\label{seg_compare}
\end{figure*}

We evaluated our method on real observational data from Telescope No. 3 of the Mini-SiTian Array at Xinglong Observatory. 
The original image size is 9576$\times$6388 pixels. 
To process these images, each was overlap-tiled into 20 sub-images, which were then resampled to 256$\times$256 pixels for segmentation using the U-Net model. 
Subsequently, we applied the LSD algorithm to the segmented images to remove non-trail patterns. 
The full process of U-Net segmentation and line detection on a single Mini-SiTian image (9576$\times$6388 pixels) takes approximately 13.78 seconds. 
This approach achieved a recall of 79.57\% and a precision of 74.56\%, with TP (true positives) = 1184, FN (false negatives) = 304, and FP (false positives) = 404. 
The formulas for recall and precision are defined as follows:
\begin{equation}
  Recall = \frac{TP}{TP + FN}
\end{equation}
\begin{equation}
  Precision = \frac{TP}{TP + FP}
\end{equation}
where TP represents the number of correctly detected trails, FP denotes false detections, and FN refers to missed trails. 
Recall reflects the model’s ability to identify all relevant targets, while precision measures the accuracy of its positive predictions. 
To suppress non-trail patterns during line detection, we imposed a minimum line length threshold in the LSD algorithm. 
This is effective because most background-induced patterns are significantly shorter than satellite trails. 
For faint trails, the U-Net model often captures only segments of the trail, resulting in fragmented outputs, as shown in Fig.\ref{seg_compare2}. 
These partial detections usually produce longer segments than background noise, though not always. 
To address this, we extend the line segments detected by LSD to span the full length of the image, effectively masking the entire trajectory of the satellite trail.

\begin{figure*}
\centering
\subfloat[Original image\label{fig:missed11}]{
     \includegraphics[width=0.45\textwidth]{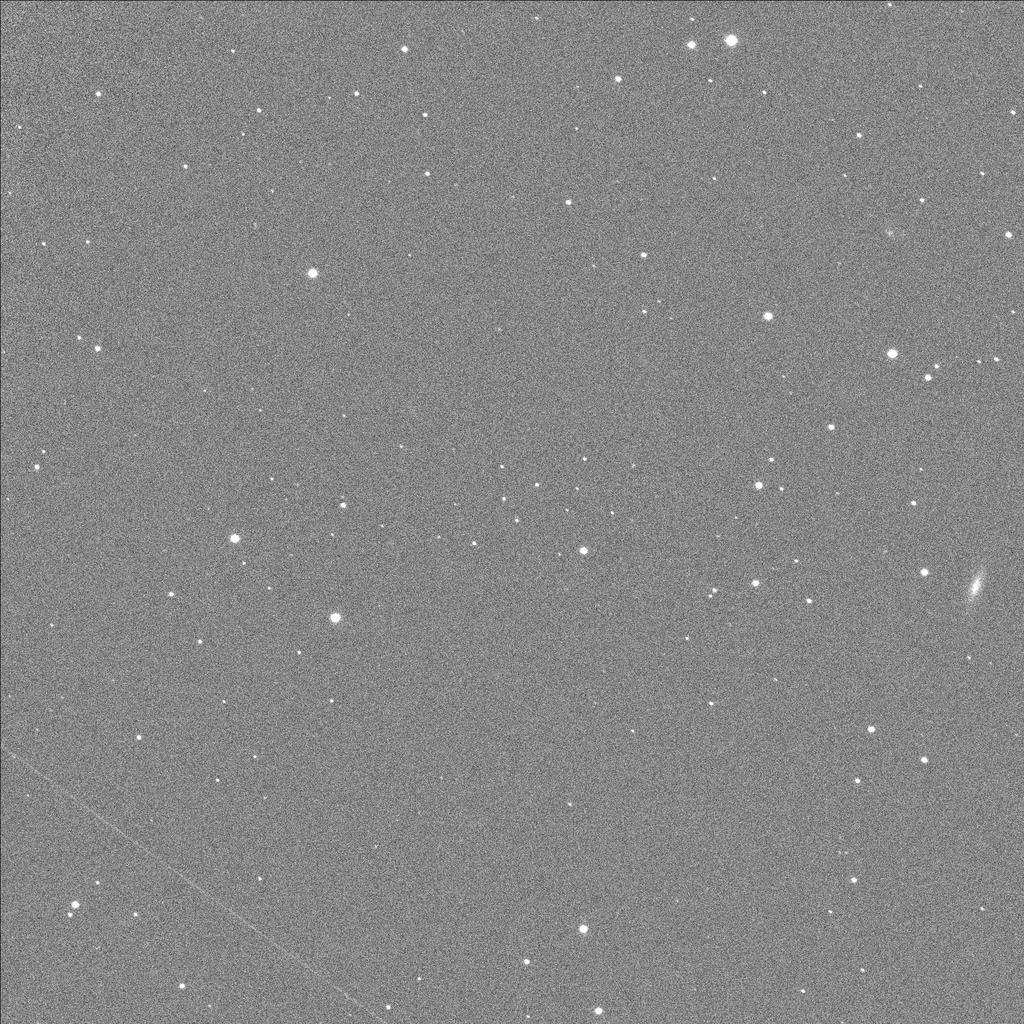}
}
\hfill
\subfloat[U-Net segmentation\label{fig:missed12}]{
     \includegraphics[width=0.45\textwidth]{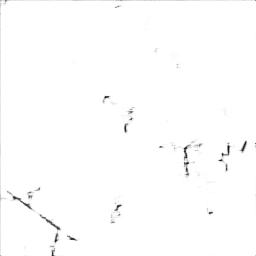}
}

\vspace{0.5cm} 

\subfloat[Original image\label{fig:missed21}]{
     \includegraphics[width=0.45\textwidth]{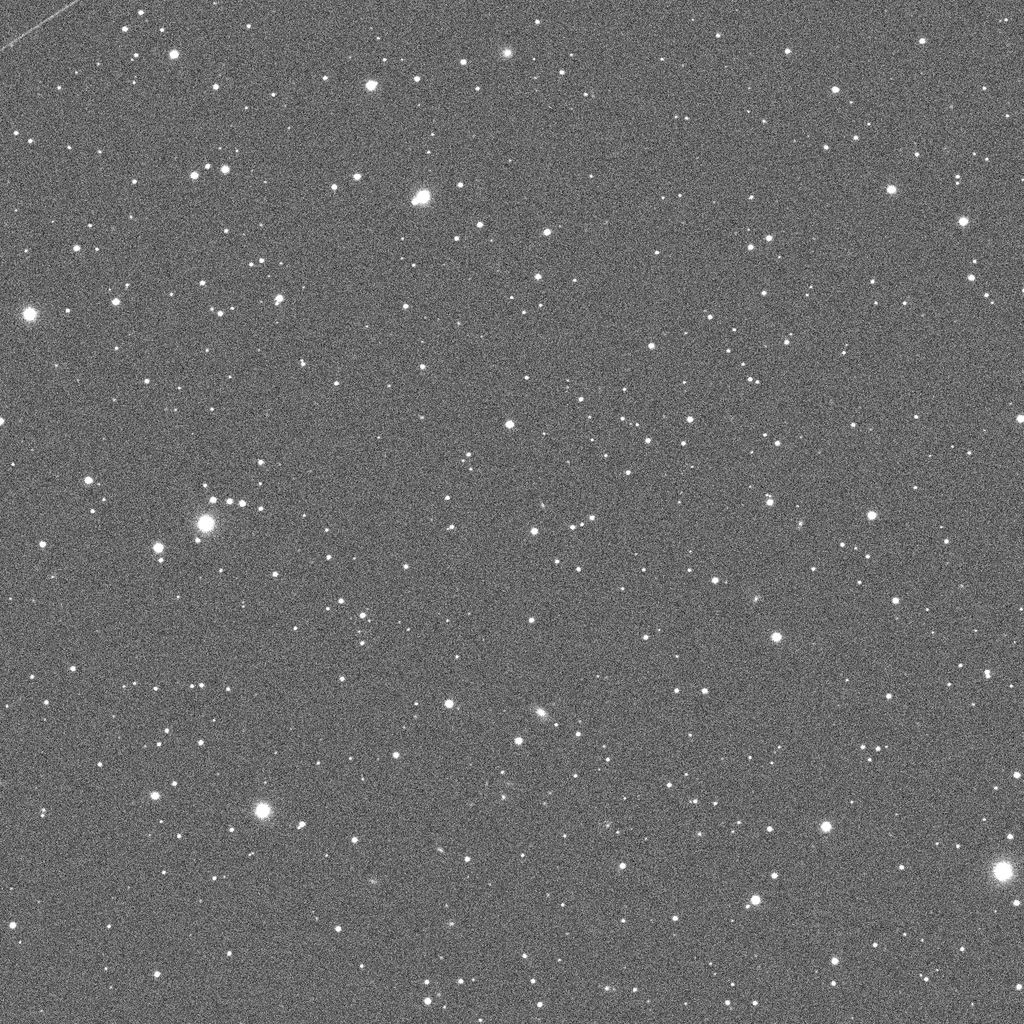}
}
\hfill
\subfloat[U-Net segmentation\label{fig:missed22}]{
     \includegraphics[width=0.45\textwidth]{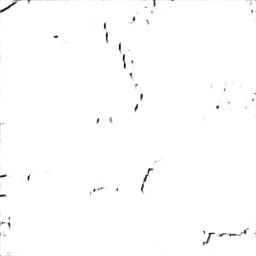}
}

\caption{\small Missed detection cases: (b) shows the U-Net segmentation map corresponding to (a), where the trail was excluded by the LSD algorithm due to incomplete segmentation. In (d), the U-Net segmentation map of (c), a very short satellite trail appears in the upper-left corner but was filtered out by LSD because its short length—caused by its position near the image edge—did not meet the detection threshold.}
\label{fig:missed}
\end{figure*}

Although our model was trained on relatively simple simulated data—featuring single trails with uniform brightness—it demonstrates strong performance 
when applied to complex real observational images. These include cases with multiple satellite trails (Fig.\ref{seg_compare1}) 
and trails exhibiting periodic brightness variations (Fig.\ref{seg_compare21}). 
This robustness can be attributed to U-Net’s inherent ability to detect elongated structures. 
Training on uniform examples allows the model to focus on linear features, enabling it to generalize well even in more complicated scenarios.
Nevertheless, some instances of missed detections persist in real observational data. These can be broadly categorized into three main types:
\begin{enumerate}
    \item Faint trails (with a SNR mostly less than 3), which the U-Net fails to detect entirely—accounting for 79.28\% of all missed detections.
    \item Trails that are partially segmented by U-Net, resulting in incomplete representations that are subsequently rejected by the LSD algorithm (e.g., Fig.\ref{fig:missed11}, Fig.\ref{fig:missed12}); these constitute 6.25\% of missed detections.
    \item Short trails located near the image corners that are filtered out due to falling below the minimum length threshold used in LSD (e.g., Fig.\ref{fig:missed21}, Fig.\ref{fig:missed22}), also comprising 6.25\% of the missed cases.
\end{enumerate}

In these cases, current line detection strategies in the U-Net output—particularly the differentiation between true trails and background artifacts—remain insufficient. 
To evaluate the impact of the minimum line length threshold in LSD, we varied this parameter and recorded the corresponding recall and precision values. The results are summarized in Table~\ref{tab:PR}.
The remaining 8.22\% of missed detections involve more complex scenarios, 
such as cluttered backgrounds, images with multiple trails of significantly varying brightness and so on.
To benchmark our approach, we compared it against two commonly used line detection methods—those of Nir \citep{2018AJ....156..229N} and the approach combining U-Net with the classical Hough transform \citep{Hough1962}—for detecting artificial satellite trails in the Mini-SiTian images. 
The results are summarized in Table~\ref{tab:PRmatrix}, and it can be observed that our method performs slightly better when dealing with complex real-world images.

\begin{table}
\centering
\caption{Test Results of Our Detection Model at Different Minimum Length Thresholds in LSD.}\label{tab:PR}
\begin{threeparttable}
\begin{tabularx}{\textwidth}{X X X X X X}
\toprule
Threshold &  TP   & FP & FN  &  Recall  & Precision\\
\midrule
0  & 1372 & 27559 & 116 & 92.20\% & 4.74\%\\
0.12  & 1184 & 404 & 304 & 79.57\% & 74.56\%\\
0.14  & 1136 & 60 & 352 & 76.34\% & 94.98\%\\
0.20  & 994 & 5 & 494 & 66.80\% & 99.50\%\\
0.39  & 735 & 2 & 753 & 49.40\% & 99.73\%\\
0.78  & 410 & 0 & 1078 & 27.55\% & 100\%\\
\bottomrule
\end{tabularx}
\begin{tablenotes}
\item Note: The threshold values in the table represent the ratio of the minimum line length in LSD to the side length of the image.
\end{tablenotes}
\end{threeparttable}
\end{table}


\begin{table}
\centering
\caption{Detection Results of Artificial Satellite Trails by Various Methods in Mini-SiTian Images.}\label{tab:PRmatrix}
\begin{threeparttable}
\begin{tabularx}{\textwidth}{p{3cm} p{2cm} p{2cm} p{2cm} p{3cm} p{2cm}}
\toprule
Method &  TP   & FP & FN  & Recall & Precision\\
\midrule
\textbf{Our method} & \textbf{1184} & \textbf{404} & \textbf{304} & \textbf{79.57\%} & \textbf{74.56\%}\\
U-Net + Hough & 902 & 204 & 586 &  60.62\% (78.46\%)\textsuperscript{*} & 81.56\%\\
Nir (2018) & 323 & 35 & 1118 &  24.87\% (77.05\%)\textsuperscript{**} & 90.46\%\\
\bottomrule
\end{tabularx}
\begin{tablenotes}
\item Note: Parenthesized recall values with asterisks (*, **) represent the estimated performance of our method at the same precision levels as the compared methods (81.56\% and 90.46\% respectively), obtained through linear interpolation of the precision-recall curve from Table~\ref{tab:PR}.
\end{tablenotes}
\end{threeparttable}
\end{table}

\section{Conclusions}
\label{sect:conclusion}
In this paper, we propose a method for detecting artificial satellite trails using a deep learning approach. 
We developed an image segmentation model based on the U-Net architecture, enhanced with classical line detection techniques. 
The model was trained on a dataset of 375 simulated satellite trail images with SNRs ranging from 2 to 30. 
On the simulated test set, the model achieved a detection rate exceeding 99\% for trails with SNRs greater than 3. 
We further validated the model on real observational data from the Mini-SiTian Array No. 3 telescope at Xinglong Observatory. 
When evaluated on these real observational images, the method—optimized with a well-tuned set of LSD parameters (minimum length threshold)—achieves a precision of 74.56\% while maintaining a recall of 79.57\%. 
These results demonstrate its effectiveness and strong generalization capability in detecting artificial satellite trails under practical observational conditions. 
This tool is currently being deployed and tested on the designated server and will eventually be used for data processing within the SiTian project.

Our model was trained exclusively on simulated data, which allows for precise control over the SNR distribution of the satellite trails. This approach has enabled the model to achieve good performance on both simulated datasets and real observational datasets.
If a more sensitive model is needed in the future, several potential refinements could be considered. For instance, as telescopes accumulate more observational data, more real data from these observations could be used for model training. This could provide additional context and variability, potentially improving the model’s robustness when applied to real-world scenarios.
Additionally, while our model can identify images containing multiple satellite trails, it may occasionally fail to detect fainter trails when there is a significant disparity in brightness between them. Incorporating training samples with multiple trails of varying brightness levels could further improve the model’s robustness and detection performance, ensuring it can reliably identify fainter trails in the presence of brighter ones.
These refinements may elevate its capabilities to an even higher level of robustness and generalization, making it better suited for practical applications in real observational conditions.

\begin{acknowledgments}
     This work was supported by the Strategic Priority Research Program of the Chinese Academy of Sciences, Grant No. XDB0550100. 
     J.L is supported by the National Natural Science Foundation of China (NSFC; grant No. 12273027).
     
     The SiTian project is a next-generation, large-scale time-domain survey designed to build an array of 60 optical telescopes, primarily located at observatory sites in China. This array will enable single-exposure observations of the entire northern sky with a cadence of only 30-minute, capturing true color (gri) time-series data down to about 21 mag. This project is proposed and led by the National Astronomical Observatories, Chinese Academy of Sciences (NAOC). As the pathfinder for the SiTian project, the Mini-Sitian project utilizes an array of three 30 cm telescopes to simulate a single node of the full SiTian array. The Mini-Sitian has begun its survey since November 2022. The SiTian and Mini-SiTian have been supported from the Strategic Pioneer Program of the Astronomy Large-Scale Scientific Facility, Chinese Academy of Sciences and the Science and Education Integration Funding of University of Chinese Academy of Sciences.
\end{acknowledgments}

\end{document}